\documentclass[10pt,twocolumn,letterpaper]{article}
\usepackage{iccv}
\usepackage{times}
\usepackage{epsfig}
\usepackage{graphicx}
\usepackage{amsmath}
\usepackage{amssymb}
\usepackage{multirow}
\usepackage{bbm}
\usepackage{commath}
\usepackage{enumitem}

\usepackage[pagebackref=true,breaklinks=true,colorlinks,bookmarks=false]{hyperref}
\usepackage[font={small}]{caption, subfig}

\iccvfinalcopy 

\def\iccvPaperID{3561} 

\ificcvfinal\fi

\begin{document}

\title{CR-Fill: Generative Image Inpainting with Auxiliary Contexutal Reconstruction}

\author{{\hfill Yu Zeng$^1$
\hfill
Zhe Lin$^2$
\hfill
Huchuan Lu$^3$
\hfill
Vishal M. Patel$^4$\hfill\hfill}\\
{\tt\small zengxianyu18@qq.com, zlin@adobe.com,  lhchuan@dlut.edu.cn, vpatel36@jhu.edu}\\
{\hfill $^1$Tencent Lightspeed \& Quantum Studios\hfill $^2$Adobe Research\hfill\hfill}\\
{\hfill $^3$Dalian University of Technology\hfill $^4$Johns Hopkins University \hfill\hfill}\\
}

\twocolumn[{%
\renewcommand\twocolumn[1][]{#1}%
\maketitle
\begin{center}
    \centering
    \includegraphics[width=.99\textwidth]{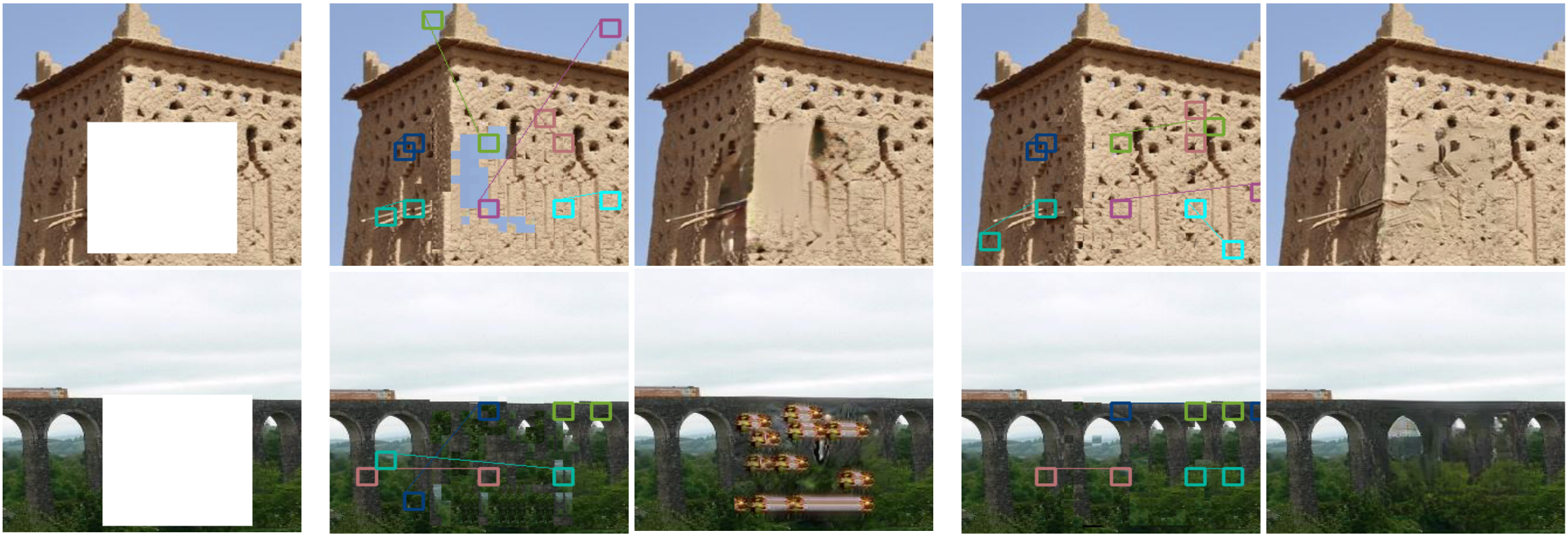}\\
\small{\hfill\hfill{(a)}  \hfill\hfill\hfill   {(b)} \hfill\hfill\hfill   {(c)} \hfill\hfill\hfill {(d)}  \hfill\hfill\hfill   {(e)}\hfill\hfill}
    \vspace{15pt}
    \captionof{figure}{(a) Input images with holes. (b) Sample patch correspondences between missing region and known region established by contextual attention layer, overlaid with the image composed by copying image patches from the known region to the corresponding position in missing region. (c) Inpainting results of DeepFillv2~\cite{yu2019free} with CA layer. (d) Patch correspondences generated by the learned CR loss. (e) Inpainting results of our attention-free generator trained with CR loss.  }
        \label{fig:intro}
\end{center}%
}]
\ificcvfinal\thispagestyle{empty}\fi



\begin{abstract}
Recent deep generative inpainting methods use attention layers to allow the generator to explicitly borrow feature patches from the known region to complete a missing region. Due to the lack of supervision signals for the correspondence between missing regions and known regions, it may fail to find proper reference features, which often leads to artifacts in the results. Also, it computes pair-wise similarity across the entire feature map during inference bringing a significant computational overhead. To address this issue, we propose to teach such patch-borrowing behavior to an attention-free generator by joint training of an auxiliary contextual reconstruction task, which encourages the generated output to be plausible even when reconstructed by surrounding regions. The auxiliary branch can be seen as a learnable loss function,~\ie named as contextual reconstruction (CR) loss, where query-reference feature similarity and reference-based reconstructor are jointly optimized with the inpainting generator. The auxiliary branch (~\ie CR loss) is required only during training, and only the inpainting generator is required during the inference. Experimental results demonstrate that the proposed inpainting model compares favourably against the state-of-the-art in terms of quantitative and visual performance. 
\end{abstract}

\section{Introduction}
\label{sec:intro}
Image inpainting is a task of predicting missing regions in images. It is an important problem in computer vision and can be used in many applications, \textit{e.g. } image restoration, compositing, manipulation, re-targeting, and image-based rendering~\cite{barnes2009patchmatch,levin2004seamless,park2017transformation}.
Traditional methods such as~\cite{efros2001image,kwatra2005texture,barnes2009patchmatch} borrow example patches from known regions or external dataset and paste them into the missing regions. They cannot hallucinate novel image contents for challenging cases involving complex, non-repetitive structures. Recent research efforts have shifted the attention to data-driven deep CNN-based  approaches~\cite{pathak2016context,iizuka2017globally,yu2018generative,liu2018image,yu2019free}. 

For inpainting, an unlimited amount of paired training data can be automatically generated simply by corrupting images deliberately and using the original images before corruption as the ground-truths. By training on large datasets, deep network-based methods have shown promising results for inpainting complex scenes. An important challenge in inpainting is that there are many plausible answers for filling in a missing region in natural images, and this ambiguity often leads to blurry or distorted structures. To overcome this issue, recent methods~\cite{yu2019free,liu2019coherent,zeng2019learning,zeng2020high,xiong2019foreground} try to reduce the uncertainty by explicitly assigning a known region as a reference for filling a missing region. This can be implemented as a patch-borrowing operation,~\eg contextual attention (CA) module~\cite{yu2018generative}, which copies feature patches from the known reference region and pastes them into the missing region.

However, there is no direct supervision on feature similarity nor the information of patch correspondences in the CA module, thus sometimes inappropriate patches are chosen with higher weights, which leads to artifacts in the consequent inpainting results. Consider the image shown in Fig.~\ref{fig:intro} (a) with missing pixels. Fig.~\ref{fig:intro} (b) shows the reference patches selected by CA layer with the largest weight. 
For visual analysis, we copy each reference patch to the corresponding location in the missing region. (c) shows the inpainting results of DeepFillv2~\cite{yu2019free}. By comparing Fig.~\ref{fig:intro} (b) and Fig.~\ref{fig:intro} (c), we can see the artifacts are caused by the incorrect reference regions found by the CA module. 
Moreover, a patch-borrowing operation in the CA layer requires computing the similarity of every pair of patches in the feature map, which is computationally expensive especially for high-resolution images. 

To address these issues, we consider to avoid using the explicit patch-borrowing in the inpainting generator (to make the generator efficient and robust to borrowing incorrect reference patches) while retaining and encouraging the patch/feature copy-pasting behaviors through jointly training of contextual reconstruction (to make the result realistic). The idea is partially motivated by Generative Adversarial Networks where its generator and discriminator are trained jointly with losses on both, while only the generator is used during testing. 

Specifically, we propose to attach an auxiliary contextual reconstruction branch to the inpainting network as a new training loss. The auxiliary branch can be regarded as a contextual reconstruction loss (CR loss) which encourages the generated output to be plausible even when reconstructed by surrounding (contextual) regions/features. CR loss not only encourages the individual features of generated missing regions reconstructable from features of the surrounding regions but also the complete contextually reconstructed composite visually plausible, in analogy to a jigsaw puzzle.

Through extensive experiments, we validate that a vanilla attention-free CNN trained with CR loss can learn to inherit the patch-borrowing behavior which was needed to be explicitly enforced with attention layers by previous approaches, as shown in Fig.~\ref{fig:intro}. Moreover, since CR loss is required only during training, there is no computational overhead brought to the inpainting model during inference, making the model much more efficient for testing. 


We summarize the contributions of this paper as follow:
\begin{itemize}[noitemsep]
\item A new learnable, auxiliary contextual reconstruction branch/loss (CR loss) to encourage the generator network to borrow appropriate known regions as references for filling in a missing region.
\item An efficient and robust attention-free inpainting generator trained jointly with the traditional inpainting loss and the auxiliary contextual reconstruction loss while allowing efficient inference for testing.
\item Extensive experiments showing the effectiveness of the CR loss and favorable performances over the state-of-the-art methods. 
\end{itemize}

\section{Related work}
Earlier inpainting methods rely on the principle of borrowing known regions to fill missing regions. Diffusion-based methods~\cite{ballester2001filling,bertalmio2000image} propagate neighboring content to the missing regions and often result in significant artifacts when filling large holes or when texture is of large variation. 
Patch-based methods~\cite{efros2001image,kwatra2005texture,barnes2009patchmatch} search for the most similar patches from known regions to complete missing regions. They can produce high-quality results for textures and repeating patterns. However, due to the lack of high-level structural understanding and inability of generating new content, the results may not be semantically reasonable. 
Encouraged by the success of deep CNNs in image restoration tasks, recent research efforts have shifted their attention to deep learning-based methods~\cite{yeh2017semantic,xie2012image,liu2018image,nazeri2019edgeconnect,li2020deepgin,li2020recurrent}. 
To produce sharper results, these methods typically adopt adversarial training inspired by GANs~\cite{goodfellow2014generative}. Pathak~\textit{et al.}~\cite{pathak2016context} first attempted to use a CNN for hole filling. 
Li~\textit{et al.}~\cite{li2017generative} propose a deep generative model for face completion. 
Iizuka~\textit{et al.}~\cite{iizuka2017globally} use two discriminators to make the inpainted content both locally and globally consistent. 

Inspired by examplar-based inpainting methods, patch-borrowing operations have been integrated into deep learning models. 
Yu~\textit{et al.}~\cite{yu2018generative} propose a contextual attention layer (CA layer) which replaces the generated features in the missing region with linear combinations of feature patches from known region using similarity as weight. 
Zeng~\textit{et al.}~\cite{zeng2019learning} propose to use region affinity from a high-level feature map to guide the patch replacement operation in the previous low-level feature map. Yang~\textit{et al.}~\cite{yang2017neuralpatch} propose a multi-scale neural patch synthesis approach based on joint optimization of image content and texture constraints. Zeng~\textit{et al.}~\cite{zeng2020high} use a related neural patch-vote approach to upsample but avoids the slow optimization by using a modified contextual attention layer. Yan~\textit{et al.}~\cite{yan2018shift} shift the encoder features of the known region to the missing region in the mirrored layer of the decoder serving as an estimation of the missing parts. Song~\textit{et al.}~\cite{song2018contextual} propose a patch-swap layer that replaces each feature patch in the missing regions with the most similar patch on the known regions. Beside these methods, related variants of patch replacement operations have also been used in other image inpainting methods \cite{liu2019coherent,liu2020rethinking,yu2019free,xiong2019foreground}. 
These methods implement a heuristic patch-borrowing operation in a network based on the distance between deep features in missing regions and known regions. Different from them, the proposed method teaches a vanilla CNN to borrow more reasonable patches by jointly training a contextual reconstruction branch. 

Previous research has studied attention-based loss function for image inpainting. Ma~\etal~\cite{ma2019coarse} propose correlation loss, which is the L1 loss between the feature affinity matrices of the inpainting result and the ground-truth. It directly penalizes the difference of similarity structure of the generated image from the ground-truth, which is too strict for large missing regions. In comparison, CR loss encourages visual realism of the reconstructed image without explicitly forcing it to match the ground-truth. This is more relaxed and provides better guidance to copy appropriate known patterns. 


\section{Generative inpainting network}
\label{sec:net}
We adopt generative adversarial network-based approach for image inpainting, which has a generator and a discriminator. The objective of the discriminator is to discriminate between real images (without any missing pixels) and images inpainted by the generator. 

\noindent \textbf{Discriminator. } We use a PatchGAN discriminator~\cite{isola2017image} with spectral normalization following \cite{yu2019free}. Inpainted images or the ground-truth complete images are passed to the discriminator, resulting in a score map, where each element is a score corresponding to a local region of the input covered by its receptive field. The loss for the discriminator is:
\begin{equation}
\begin{split}
\mathcal{L}_D =& \mathbb{E}_{X\sim p_{data}(X)}\left[\mbox{ReLU}(\mathbbm{1}-D(X))\right] +\\ &\mathbb{E}_{U\sim p_{U}(U)}\left[\mbox{ReLU}(\mathbbm{1}+D(G(U)\circ M+U))\right],  
\end{split}
\end{equation}
where $D$ denotes the discriminator, $X$ represents the real image (ground-truth), $U$ represents the incomplete image with the pixels in the missing regions set to zero, $M$ represents a binary mask corresponding of the missing region where $M_{xy}=1$ indicates that pixel at $x,y$ is missing and $M_{xy}=0$ indicates that pixel at $x,y$ is valid/known, $G(\cdot)$ represents the generator and $\circ$ denotes element-wise multiplication. The inpainting result $G(U)\circ M+U$ is composed by putting the generator $G(U)$ in the missing region and keeping the original content of $U$ in the known region. 

\begin{figure}[h]
\begin{center}
\includegraphics[width=\linewidth]{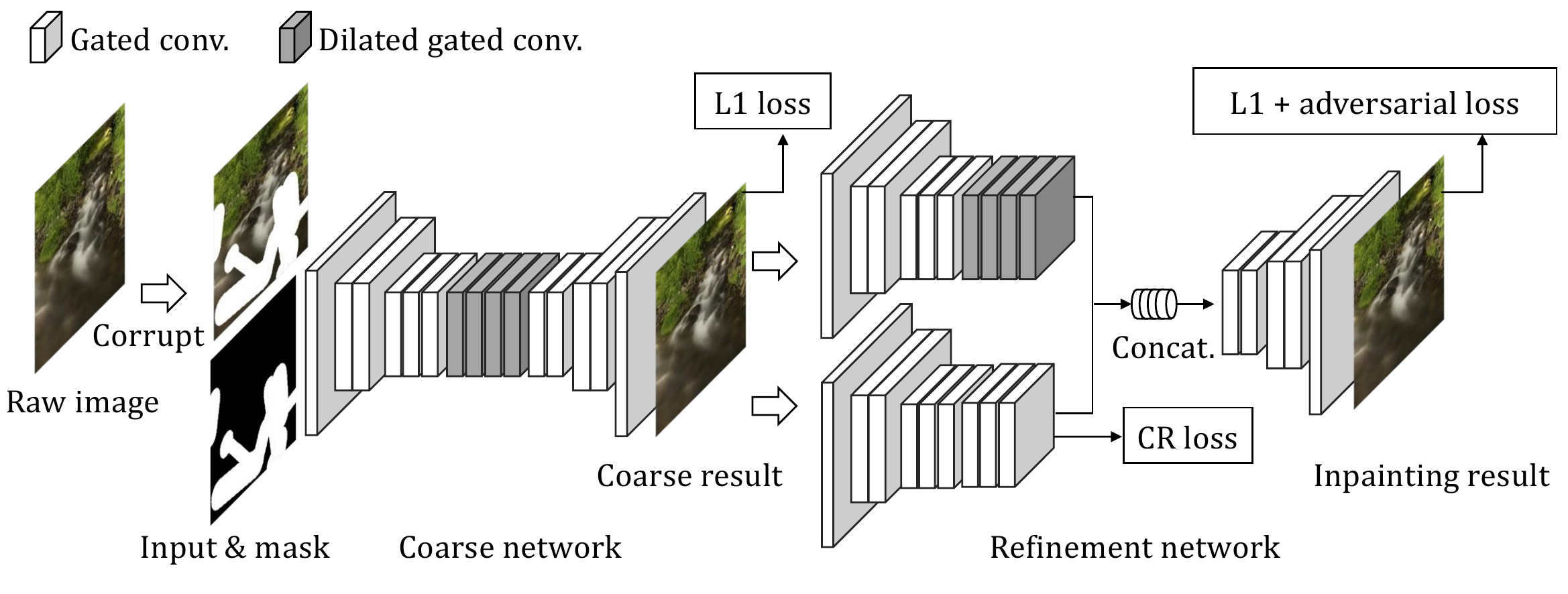}\\
\end{center}
   \caption{Overall architecture of the generator network. }
\label{fig:overall}
\end{figure}

\noindent \textbf{Coarse-to-fine generator.} Fig.~\ref{fig:overall} shows the overall architecture of our generator network. It is a coarse-to-fine architecture, similar to the one in DeepFillv2~\cite{yu2019free} but the CA layer is removed and CR loss is applied instead. The coarse network and the refinement network are convolutional encoder-decoder type networks. Dilated convolution layers are used to enlarge the receptive fields. We use gated convolution~\cite{yu2019free} in all convolution and dilated convolution layers. The coarse network takes an incomplete image where missing pixels are set to zero and a binary mask indicating the missing region as input and generates an initial prediction. Then the refinement network takes this initial prediction as input and outputs the final inpainting result. 

\noindent \textbf{Training the generator. } We expect the coarse network to complete the global structure and then details will be filled in by the refinement network. So only the L1 loss is used to train the coarse network.  A combination of L1 loss, adversarial loss and the proposed CR loss is used to train the refinement network.  Let $Y=G(U)$ represent the refinement network output.  Then the loss for the refinement network is defined as follow:
\begin{equation}
\label{eq_image_loss}
\mathcal{L}_{G} = \mathbb{E}_{U,X\sim p(U,X)}[L(Y)+\lambda L_{CR}], 
\end{equation}
where $L(Y)$ is the sum of L1 loss and  adversarial loss 
\begin{equation}
\label{eq:inpaint_loss}
L(Y) = \mbox{ReLU}(\mathbbm{1}-D(Y\circ M+U)) + \beta \norm{Y-X}_1. 
\end{equation}
Here, $L_{CR}$ denotes CR loss, which will be elaborated in Section~\ref{sec_loss}. $\lambda$ is set equal to 0.5. $\beta$ is set equal to 1.5. 

\section{Contextual reconstruction}
\label{sec_loss}
\noindent \textbf{Revisit contextual attention.} As we have discussed in Sec.~\ref{sec:intro}, for a generator with attention-based patch-borrowing operations, inappropriately borrowed patches will lead to artifacts in the inpainted image. The artifacts are produced based on the features from the reference regions, and thus resemble their appearance in image space. 
It also can be expected that moving these patches in the missing region will not result in a reasonable image, as has been shown in Fig.~\ref{fig:intro}~(b).
On the other hand, a visually realistic image can be reconstructed with contextual patches if the reference regions are chosen correctly. 

\begin{figure}[h]
\begin{center}
\includegraphics[width=\linewidth]{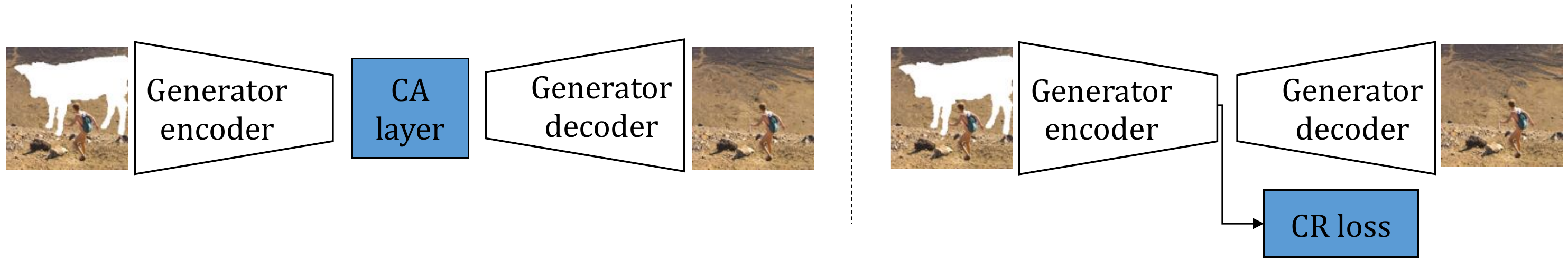}\\
\end{center}
   \caption{Comparison of the usage of CA layer and CR loss. }
\label{fig:cmp_rv}
\end{figure}

Inspired by this observation, we propose a CR loss to encourage a network to find optimal reference regions by minimizing the L1 and adversarial loss of an auxiliary result composed by image patches in known region, ~\ie as Fig.~\ref{fig:intro}~(b) and~(d) was produced but is made differentiable by exploiting softmax and an autoencoder-like auxiliary network. Furthermore, unlike previous approaches which integrate a patch-borrowing operation in the generator, CR loss distills the information of the optimal contextual patches into an attention-free generator. As illustrated in Fig.~\ref{fig:cmp_rv}, it does not directly involve the generation of inpainted images and affects the network only during training for learning better features. 

\begin{figure}[h]
\begin{center}
\includegraphics[width=\linewidth]{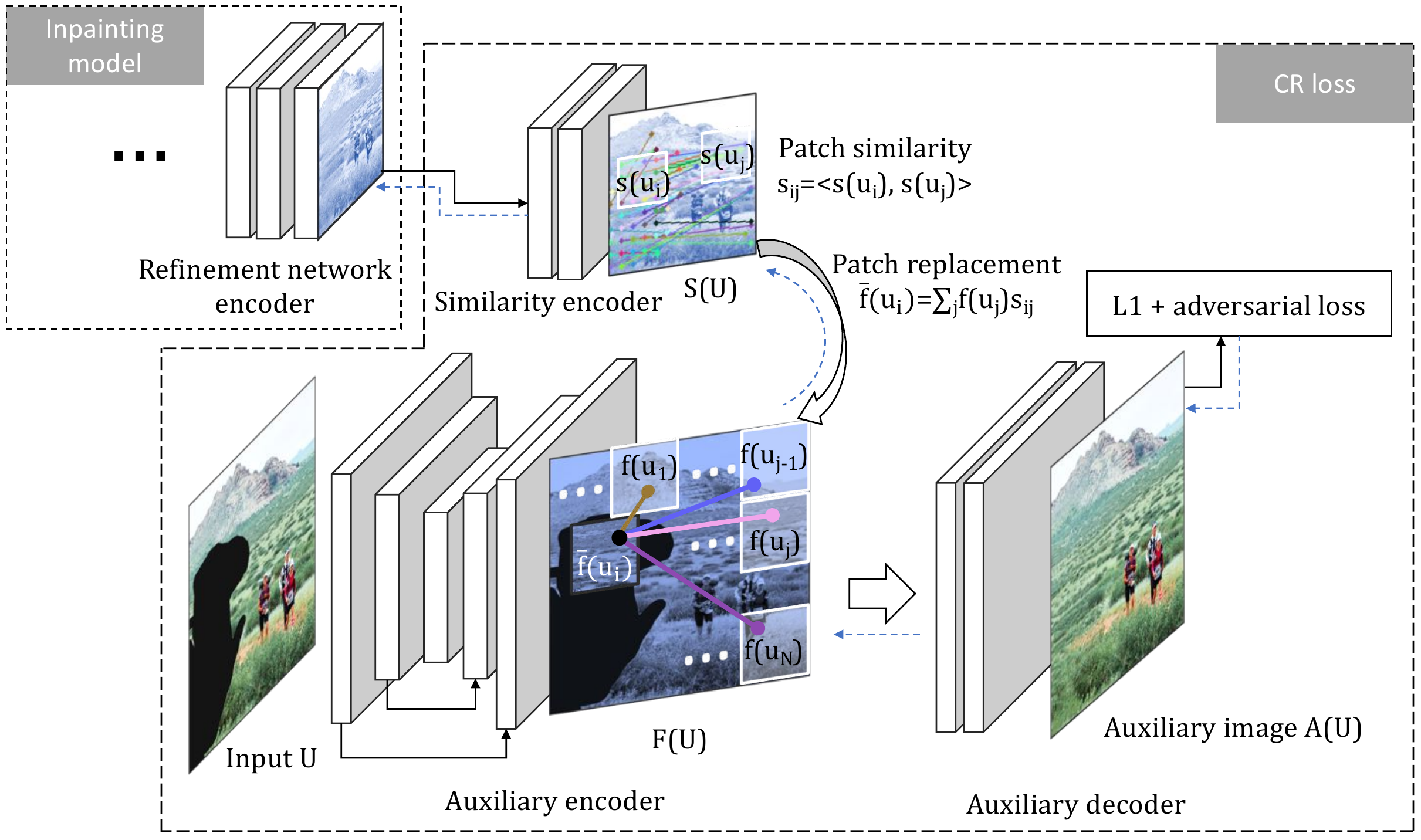}\\
\end{center}
   \caption{Training system with the proposed contextual reconstruction loss. The dashed blue lines indicates the gradient backpropagation flow. }
\label{fig:train_with_cr_loss}
\end{figure}

\noindent \textbf{Contextual reconstruction loss.} 
The overall training system with CR loss is shown in Fig.~\ref{fig:train_with_cr_loss}, where the inpainting model takes an incomplete image and the binary mask of the missing regions as input and outputs the inpainted image. The training system with CR loss consists of a similarity encoder and an auxiliary encoder-decoder type network and re-uses the inpainting loss defined in Sec.~\ref{sec:net}. The similarity encoder takes the generator feature as input and encodes the similarity among image regions. The auxiliary encoder-decoder network produce an auxiliary image in which the known regions are unchanged while the missing regions are filled with similar known regions based on the similarity provided by the similarity encoder. CR loss of the generator feature is defined to be the inpainting loss (~\textit{i.~e.~}L1 and adversarial loss) of the auxiliary image. By minimizing CR loss, the generator features are encouraged to be close to the known image features of the smallest inpainting loss. 
In what follows, we elaborate on the formal definition and explanation for this loss. 

For the convenience of the description of the proposed method, here we define the global and patch-wise perspective of a CNN. If we view an image $U$ passed through a CNN as the combination of square patches $u_1,u_2,...$, a convolution layer feature map $F(U)$ of this image $U$ can be seen as the combination of local function of image patches $f(u_1),f(u_2),...$. Hereafter an upper case letter, \textit{e.~g.~}$F(U)$, will represent a whole image or feature map with the corresponding lower case one for a patch in it \textit{e.~g.~}$f(u_i)$. Each image patch can correspond to feature patches of various sizes in different layers of a CNN. 
\begin{figure}[h]
\begin{center}
\includegraphics[width=.7\linewidth]{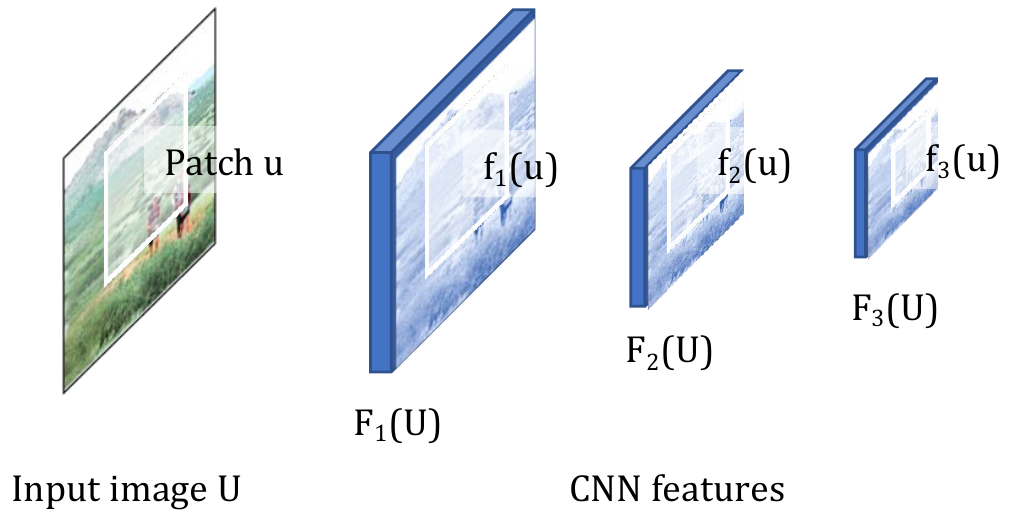}\\
\end{center}
   \caption{If view an input image as combination of patches, corresponding feature patch in a layer $l$ can be seen as a representation $f_l(u)$ of an image patch $u$. }
\label{fig:def_patch}
\end{figure}
For example, for the layers shown in Fig.~\ref{fig:def_patch}, an $N\times N$ image patch $u$ centered at $(x,y)$ corresponds to an $\frac{N}{2}\times\frac{N}{2}$ feature patch $f_2(u)$ centered at $(\frac{x}{2}, \frac{y}{2})$ in the second layer and an $\frac{N}{4}\times\frac{N}{4}$ feature patch $f_3(u)$ centered at $(\frac{x}{4}, \frac{y}{4})$ in the third layer. They can be seen as different representations of the same patch. Therefore, we can take the similarity of a pair of feature patches $f_l(u_i), f_l(u_j)$ at a layer $l$ and use it as a similarity $s_{ij}$ of the corresponding image patch or feature patches at arbitrary other layers. 
We take the pair-wise cosine similarity of feature patches $<s(u_i),s(u_j)>$ at the last layer of the similarity encoder as $s_{ij}$:
\begin{equation}
s_{ij} = \frac{s(u_i)^\intercal s(u_j)}{\| s(u_i)\|\cdot \|s(u_j)\|},
\end{equation}
where the feature patches $s(u_i), s(u_j)$ are viewed as vectors when taken the inner product. It can be implemented as processing the feature map with patches extracted from itself as convolution filters as described in~\cite{yu2018generative}.

Before the feature of the auxiliary encoder is passed through the auxiliary decoder, each feature patch is replaced with a weighted sum of patches in the known region with the softmax of the similarity provided by the similarity encoder as weight. Let $f(u_i)$ be the auxiliary encoder feature of image patch $u_i$,  $\bar{f}(u_i)$ the feature of $u_i$ after patch replacement is obtained as follows,
\begin{equation}
\label{eq:replace}
\bar{f}(u_i) = \sum_{j \in \mathcal{V}} \mbox{softmax}(\alpha s_{ij}) f(u_j), 
\end{equation}
where $\mathcal{V}$ represents the index set of patches in the known region. $\alpha$ is set equal to 10. It can be implemented as transposed convolution which processes the similarity maps with patches extracted from the auxiliary encoder features as filters after dropout with the similarity to all patches in the missing region. 

Then the feature map after the patch replacement is translated into an auxiliary image by the auxiliary decoder. Let $\bar{F}(U)$ be the feature map consisting  of feature patches $\bar{f}(u_i)$ given in Eqn.~\ref{eq:replace}, the auxiliary image $Aux(U)$ of the input image $U$ is obtained as follow,
\begin{equation}
Aux(U) = H(\bar{F}(U)),
\end{equation}
where $H()$ represents the auxiliary decoder connected to the auxiliary encoder. Based on this discussion, the contextual reconstruction loss $L_{CR}$ is defined as the inpainting loss $L()$ (Eqn.~\ref{eq:inpaint_loss}) of the auxiliary image:
\begin{equation}
L_{CR} = L(Aux(U)). 
\end{equation}

\begin{figure}[h]
\begin{center}
\includegraphics[width=\linewidth]{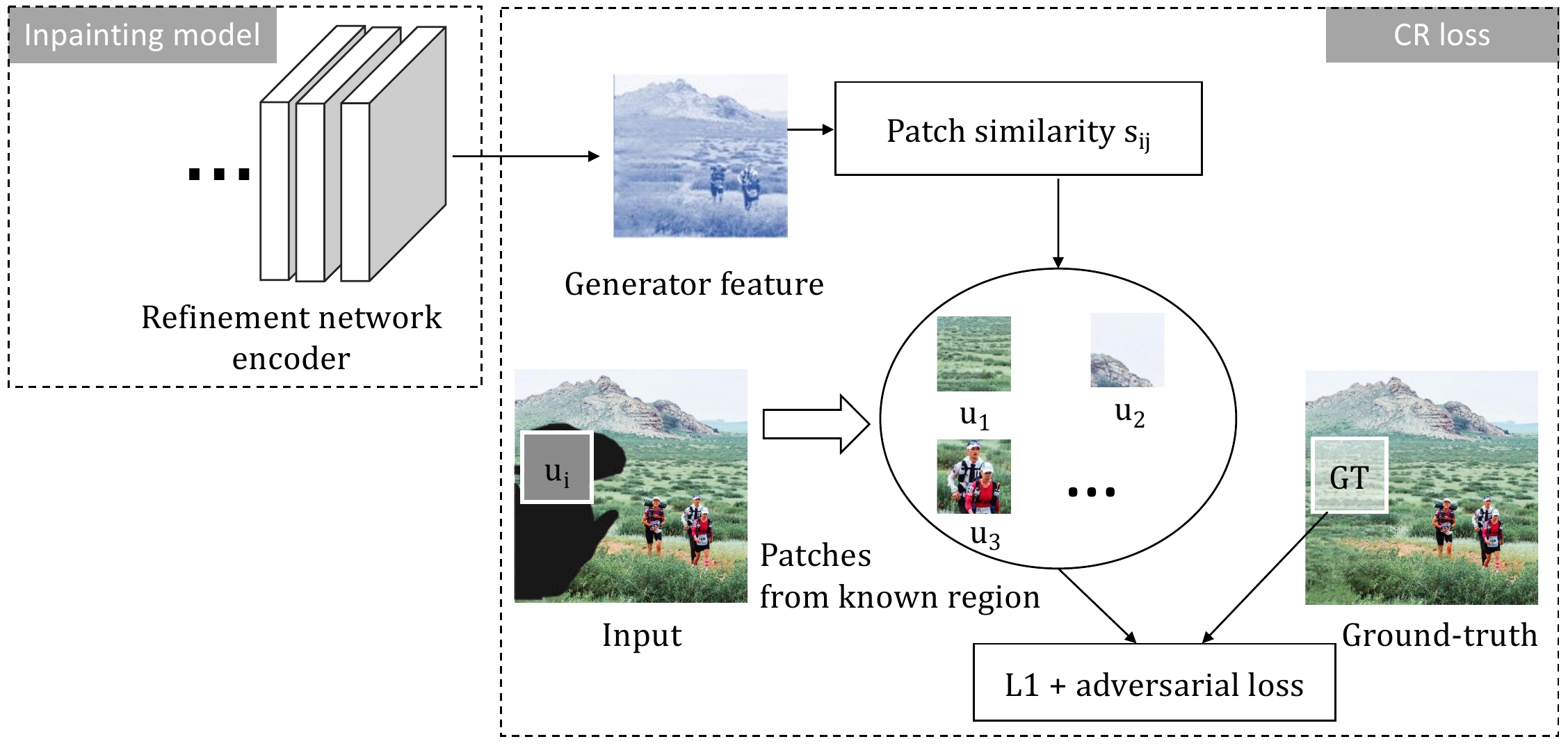}\\
\end{center}
   \caption{A simplified view of the training system.}
\label{fig:loss_simp}
\end{figure}

Assumes that the auxiliary decoder inverts the auxiliary decoder for known regions. Then the auxiliary encoder-decoder approximately conducts the process of producing Fig.~\ref{fig:intro} (b) and~(d) described in Sec.~\ref{sec:intro}: for each patch $u_i$ in the missing regions, the most similar patch $u_{i^*}$ from the known regions, \textit{i.~e.~} $i^*=\arg\max_j s_{ij}$, is moved to the position of $u_i$. 
This results in a simplified view of the training system shown in Fig.~\ref{fig:loss_simp}, which is an analogue of learning to solve a jigsaw puzzle. To minimize the loss of the auxiliary image, proper known patches should be chosen and moved to the right place. Furthermore, as the reference known patches are selected according to the similarity of generator features, it requires the generator feature in the missing region to be closest to the proper known region. Since the searching of known patches is throughout the whole image, distant relationship can be captured. A more detailed explanation will be given in the following section. 

\noindent \textbf{Explanation. }For a patch in the known region, \textit{i.~e.~}$u_j, j \in \mathcal{V}$, since $\bar{f}(u_j)\approx f(u_j)$ as $\mbox{softmax}(\alpha s_{jj})\approx 1$, the auxiliary image patch in the known region will be $a(u_j)=h(f(u_j)), j\in \mathcal{V}$. Note that the ground-truth of $a(u_j)$ is $u_j$ itself for $j\in \mathcal{V}$, the auxiliary encoder-decoder $H(F(U))$ will learn lazily to copy the input to the output so it is easy to make the auxiliary decoder invert the auxiliary encoder: $h(f(u_j)) = u_j~\mbox{for}~j\in \mathcal{V}.$
To make it easier, we put skip connection from shallow layers to the deeper layers of the auxiliary encoder, as indicated in Fig.~\ref{fig:train_with_cr_loss}. 

Assume that softmax in Eqn.~\ref{eq:replace} to be ``hard", then the feature patch after replacement can be approximated as:
\begin{equation}
\bar{f}(u_i) = f(u_{i^*})~\mbox{where}~i^* =\arg\max_{j\in \mathcal{V}} s_{ij}.
\end{equation}
Then a patch in the auxiliary image is:
\begin{equation}
a(u_i) = h(f(u_{i^*})) = u_{i^*}~\mbox{where}~i^* =\arg\max_{j\in \mathcal{V}} s_{ij}.
\end{equation}
Resembling the translation between the global and patch-wise perspective of a CNN introduced earlier, the inpainting loss for the whole image also can be distributed to each patch. Thereby the inpainting loss of the auxiliary image $L(Aux(U))$ can be seen as a sum over local loss of patches: $L(Aux(U))=\sum_i l_i(a(u_i))$ (We differentiate the local loss of different patches as the ground-truth for each patch is different, which means $L(Aux(U))$ is in fact $L(Aux(U);X)=\sum_i l(a(u_i);x_i)$ and it is denoted as $l_i(a(u_i))$ for short). We have argued earlier that the loss on known regions can be minimized with the auxiliary encoder-decoder copying the input to the output. So what count for $L(Aux(U))$ are the patches in the missing region:
\begin{equation}
L(Aux(U)) = \sum_{i \in \mathcal{V}^C} l_i(u_{i^*}) ~\mbox{where}~i^* =\arg\max_{j\in \mathcal{V}} s_{ij}. 
\end{equation}
This is a sum over local inpainting loss of known image patches with the largest similarity to the generated patches in the missing region. 
Note that the number of known patches in an image is limited, so for each local inpainting loss $l_i()$ there must be one patch $u_{i^0}$ that has the smallest loss among all the known patches, \textit{i.~e.~}$i^0=\arg\min_{j\in\mathcal{V}} l_i(u_j)$. The minimum of CR loss is achieved when $i^* = i^0$. In other words, when the generator features in missing regions are closest to the features in known regions which have the smallest inpainting loss. 

In summary, to minimize the CR loss, two conditions are required. First, appropriate reference patches are selected from the known region.  Second, feature patches in missing regions are closest to the corresponding reference patches. Compared with CA layer which searches the known patches using heuristics with no guarantee on the reference patches it reaches, CR loss can find more reasonable reference patches. This will be supported by quantitative experimental results in Sec.~\ref{sec:ref}. 

\section{Experimental Results}
\label{sec:ref}
We implement our method with Python and PyTorch. Detailed network architectures and code can be found in the supplementary material. We train the models using Adam~\cite{kingma2014adam} optimizer with the learning rate of 0.0001. To show the effectiveness of the CR loss, we train the network shown in Fig.~\ref{fig:overall} with and without the CR loss on the Places2 training set. We add images from the salient object segmentation dataset~\cite{xiong2019foreground} following~\cite{zeng2020high,xiong2019foreground}. Square masks, irregular masks~\cite{yu2019free} and object-shaped masks~\cite{zeng2020high} are randomly switched at every mini-batch to be used for creating missing regions. Training samples are cropped to $256\times 256$ and missing regions are put at random positions. The obtained models are denoted as Baseline and Baseline+CR, respectively. 
To compare the proposed CR loss with CA layer, we also train a network with the same architecture as DeepFillv2~\cite{yu2019free} but remove the CA layer and apply CR loss instead. The training protocol for this network is made to be the same as the official implementation of DeepFillv2~\cite{yu2019free}, where Places2 training set with square and irregular masks are used for training; $\beta$ in Eqn.~\ref{eq:inpaint_loss} is set equal to 1 as in the official implementation of DeepFillv2. The obtained model is denoted as DeepFillv2-CA+CR. All evaluations are conducted on the same platform: an Ubuntu machine with a 3GHz Intel i7-9700F CPU, 32GB memory, 256GB swap space, and a GPU NVIDIA RTX2080Super with 8GB GPU memory. 
\subsection{Comparison with state-of-the-art}
We compare our methods with the following state-of-the-art methods: PENNet~\cite{zeng2019learning}, DeepFillv2~\cite{yu2019free}, Rethink~\cite{liu2020rethinking}, HiFill~\cite{yi2020contextual}. We use their official implementation and the models trained on Places2 training set provided by the authors of corresponding papers. 
We use L1 error, PSNR and SSIM to measure the performance quantitatively. For comparison on visual quality, we show the inpainting results and the results of a subject evaluation conducted by human raters. 

\noindent \textbf{Quantitative evaluation. }
Table~\ref{table_metric} and Table~\ref{table_metric2} show quantitative comparisons of our method with state-of-the-art methods on Places2 validation set containing 36500 images and ImageNet validation set containing 50000 images. All images are cropped to $256\times 256$ with missing regions of different shapes ($128\times128$ squares, object and irregular shapes) at random positions. From these table, we can see that our method has a smaller L1 error, larger PSNR and SSIM than the existing methods in most cases, which indicates that the inpainting results corresponding to our method are closer to the ground-truth. 
\begin{figure*}[t]
\begin{center}
\includegraphics[width=\linewidth]{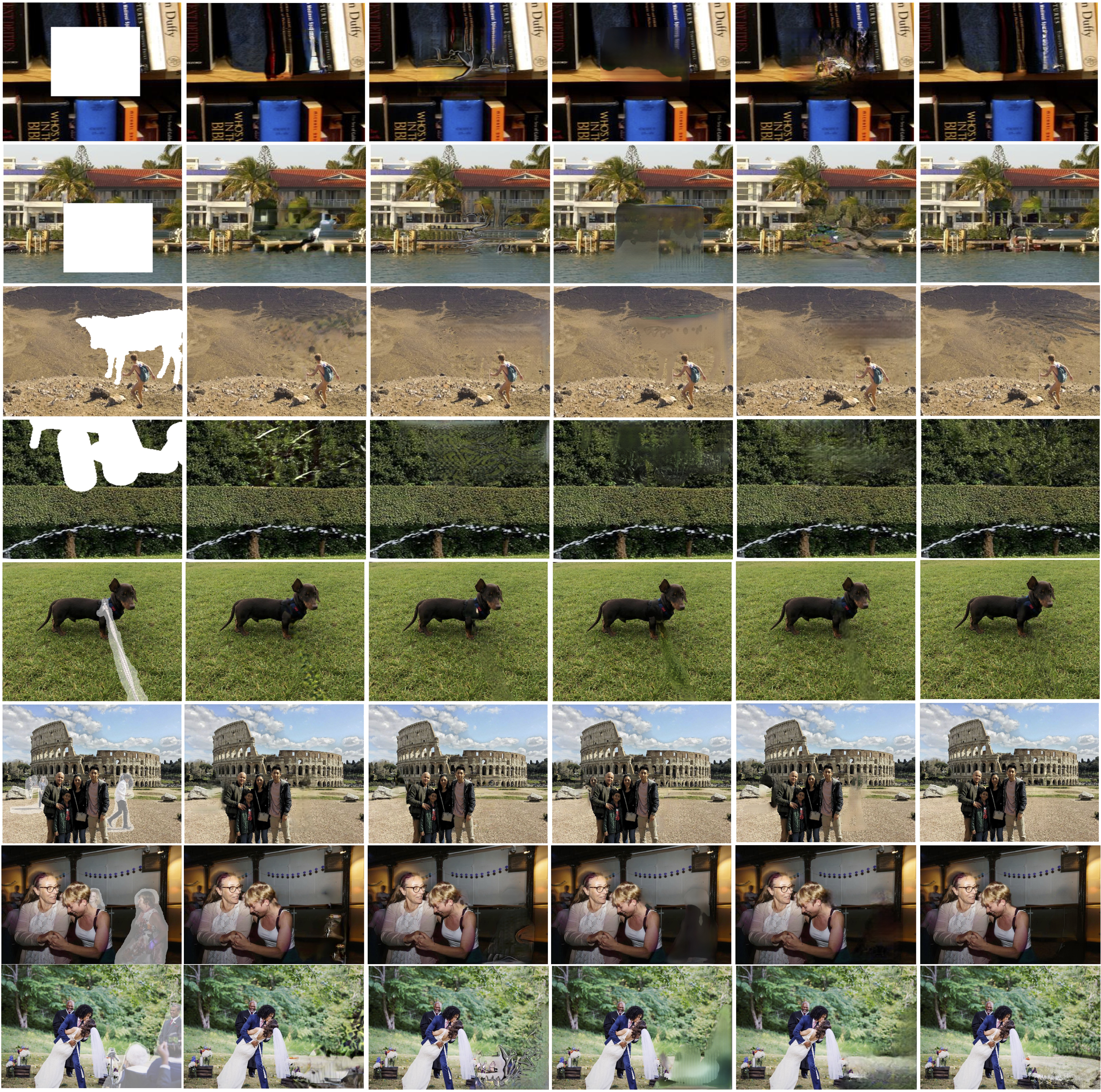}\\
\small{\hfill{Input} \hfill\hfill  {DeepFillv2} \hfill\hfill  {HiFill}\hfill\hfill{PENNet} \hfill\hfill  {Rethink} \hfill\hfill  {Ours}\hfill\hfill}
\end{center}
  \caption{Visual comparison of inpainting results of our method and other methods. Zoom-in to see the details. Images are compressed due to file size limitation. More results and comparison at high-resolution can be found in the supplementary material. }
\label{fig:visual}
\end{figure*}

\begin{table*}[t]
\setlength{\tabcolsep}{3pt}
\caption{\small Quantitative evaluation results on the Places2 validation set. The best scores are in bold.}
\label{table_metric}
\small
\begin{center}
\begin{tabular}{ccc|cccc|cccc|cccc|c}
\hline
\multicolumn{3}{c|}{\multirow{2}{*}{Method}} & \multicolumn{4}{c|}{Square holes}& \multicolumn{4}{c|}{Irregular holes}& \multicolumn{4}{c|}{Object holes}& User\\
\multicolumn{3}{c|}{}&FID & L1 error &PSNR &SSIM&FID&L1 error &PSNR &SSIM &FID& L1 error &PSNR &SSIM  &preference\\
\hline
\multicolumn{3}{c|}{DeepFillv2~\cite{yu2019free}\tiny{ICCV'19}} & 3.313 & 0.0290 & 22.72 & 0.8429 &1.387& 0.0178 & 27.03 & 0.8947 &1.905& 0.0225 & 27.95 & 0.8887 & 88\\
\multicolumn{3}{c|}{Rethink~\cite{liu2020rethinking}\tiny{ECCV'20}} &11.03& 0.0307 & 22.30 & 0.8305 &6.877& 0.0194 & 26.13 & 0.8796 & 5.539&0.0237 & 27.07 & 0.8829 &13 \\
\multicolumn{3}{c|}{PENNet~\cite{zeng2019learning}\tiny{ICCV'19}} &14.67& 0.0284 & 23.28 & 0.8402 &10.02& 0.0227 & 25.35 & 0.8699 &5.539&0.0251 & 26.28 & 0.8780 & 6\\
\multicolumn{3}{c|}{HiFill~\cite{yi2020contextual}\tiny{CVPR'20}} &18.84& 0.0329 & 21.35 & 0.8103 &6.887&0.0210 & 25.48 & 0.8712 &7.819&0.0267 & 26.14 & 0.8647 & 27\\
\multicolumn{3}{c|}{Ours} &\textbf{2.908}& \textbf{0.0263} & \textbf{23.36} & \textbf{0.8462} &\textbf{1.231}& \textbf{0.0157} & \textbf{27.79} & \textbf{0.9002} &\textbf{1.724}&\textbf{0.0202} & \textbf{28.79} & \textbf{0.8929} &\textbf{271}\\
\hline
\end{tabular}
\vspace{-10pt}
\end{center}
\end{table*}
\begin{table*}[t]
\setlength{\tabcolsep}{4pt}
\caption{\small Quantitative evaluation results on the ImageNet validation set. The best scores are in bold. }
\label{table_metric2}
\small
\begin{center}
\begin{tabular}{ccc|cccc|cccc|cccc}
\hline
\multicolumn{3}{c|}{\multirow{2}{*}{Method}} & \multicolumn{4}{c|}{Square holes}& \multicolumn{4}{c|}{Irregular holes}& \multicolumn{4}{c}{Object holes}\\
\multicolumn{3}{c|}{}& FID&L1 error &PSNR &SSIM& FID&L1 error &PSNR &SSIM & FID&L1 error &PSNR &SSIM \\
\hline
\multicolumn{3}{c|}{DeepFillv2~\cite{yu2019free}\tiny{ICCV'19}} &6.739& 0.0335 & 21.18 & 0.8200 & 2.228&0.0205 & 25.55 & 0.8765 & 3.257&0.0250 & 26.51 & 0.8739 \\
\multicolumn{3}{c|}{Rethink~\cite{liu2020rethinking}\tiny{ECCV'20}} & 13.49&0.0347 & 21.13 & 0.8137 & 6.604&0.0216 & 25.15 & 0.8682 &5.778&0.0259 & 26.11 & 0.8723 \\
\multicolumn{3}{c|}{PENNet~\cite{zeng2019learning}\tiny{ICCV'19}} & 14.67&0.0316 & \textbf{22.09} & 0.8257 & 7.462&0.0243 & 24.53 & 0.8615 &6.129&0.0269 & 25.48 & 0.8698 \\
\multicolumn{3}{c|}{HiFill~\cite{yi2020contextual}\tiny{CVPR'20}} & 16.92&0.0329 & 21.35 & 0.8103 &5.519&0.0210 & 25.48 & 0.8712 &6.611&0.0287 & 25.24 & 0.8570 \\
\multicolumn{3}{c|}{Ours} & \textbf{5.813}&\textbf{0.0302} & 21.91 & \textbf{0.8269} & \textbf{1.707}&\textbf{0.0180} & \textbf{26.43} & \textbf{0.8855} & \textbf{3.047}&\textbf{0.0225} & \textbf{27.45} & \textbf{0.8808} \\
\hline
\end{tabular}
\vspace{-10pt}
\end{center}
\end{table*}

\noindent \textbf{Visual quality. }
Fig.~\ref{fig:visual} shows the visual comparison of our method and existing methods. We can see from the figure that the missing region recovered by our method is more visually coherent with surrounding known regions. This figure implies the effectiveness of CR loss in exploiting structured information in known regions. Furthermore, unlike a previous method PENNet which has high performance in quantitative evaluation but tends to generate blurry images, the results of our method are more visually realistic. 

\noindent \textbf{User study. }
We invite 9 human raters to conduct a user study on 45 images randomly sampled from the Places2 validation set. Every $1/3$ of them are corrupted with missing regions of different shapes,~\textit{i.~e.~} square, irregular and object shaped, at random positions. Each time the incomplete image and the results by different methods are presented to the raters in random order. The raters are asked to select one best result. The number of user preference for all methods are shown in the last column of Table~\ref{table_metric}. From the user study results, we can see that the results of our method are preferred by the human raters most frequently. 

\begin{figure*}[t]
\begin{center}
\includegraphics[width=\linewidth]{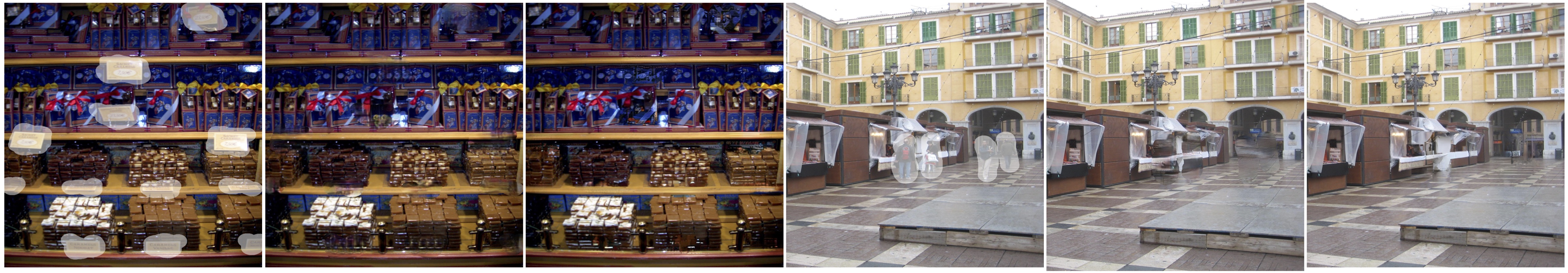}\\
\small{\hfill{Input} \hfill\hfill  {HiFill} \hfill\hfill  {Ours}\hfill\hfill{Input} \hfill\hfill  {HiFill} \hfill\hfill  {Ours}\hfill\hfill}
\end{center}
  \caption{High-resolution results ($1200\times 1600, 1152\times1536$) compared with HiFill. Zoom in to see the details. Images are compressed due to file size limitation. }
\label{fig:visual_hr}
\end{figure*}

\noindent \textbf{High-resolution inpainting. }
The efficiency of the proposed method makes it potentially capable of processing high-resolution input. We make a simple adjustment to the network to adapt to high-resolution input: running the coarse network at $1/2$ the input size and interpolate its output back to the original resolution before passed through the refinement network. Fig.~\ref{fig:visual_hr} shows high-resolution inpainting results by our adjusted inpainting model and HiFill~\cite{yi2020contextual} which was proposed for inpainting at high-resolution. More results and comparison at high-resolution can be found in the supplementary material.

\subsection{Ablation study}
\begin{figure}[h]
\begin{center}
\includegraphics[width=\linewidth]{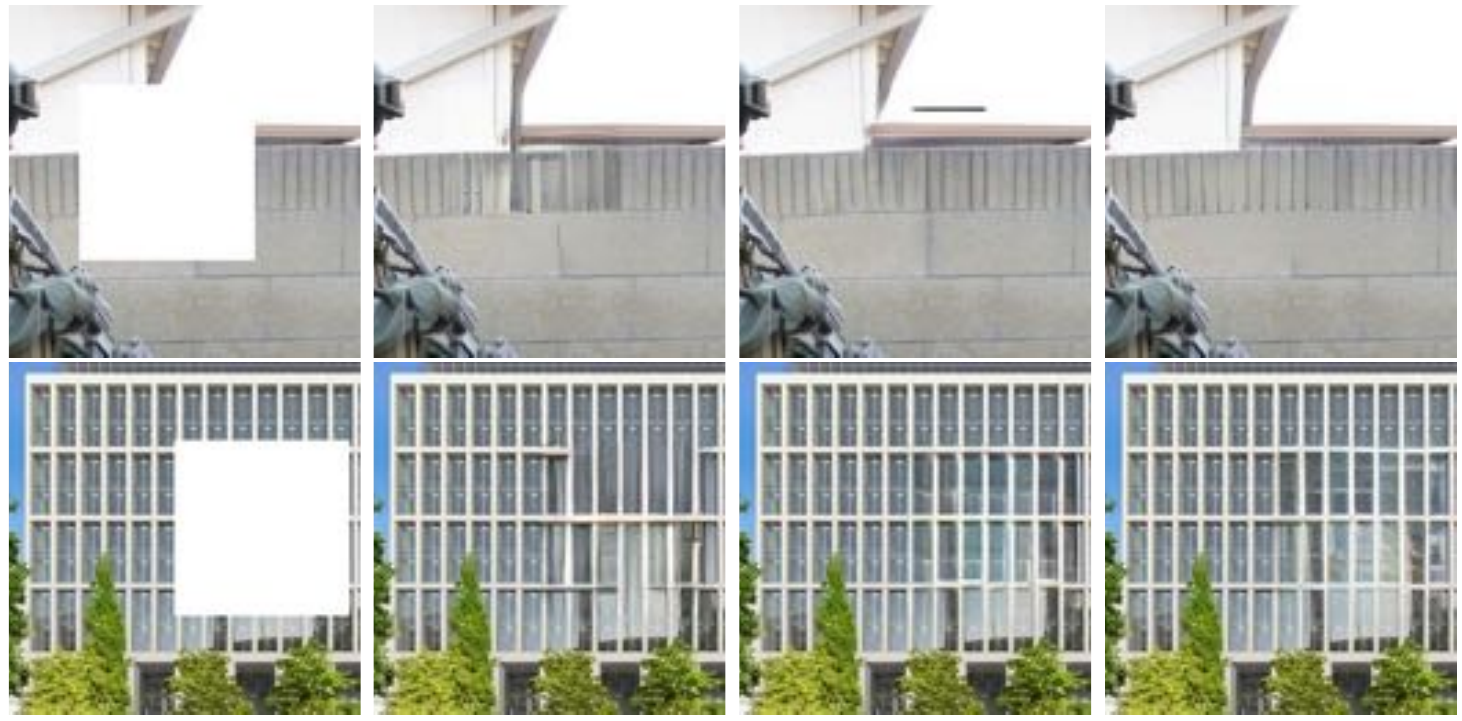}\\
\small{\hfill\hfill{Input} \hfill\hfill  {Baseline} \hfill\hfill  {CA layer} \hfill\hfill  {CR loss}\hfill\hfill\hfill}
\end{center}
   \caption{Effect of CA layer and CR loss. Both encourage to capture distant relation among image regions.}
\label{fig:abla1}
\end{figure}

\noindent\textbf{Inpainting results. }Fig.~\ref{fig:abla1} shows a visual comparison of the results obtained by the baseline model (Baseline), the baseline model trained with the CR loss (Baseline+CR), and the results of DeepFillv2 with CA layer. It can be seen that both CA layer and CR loss enable the inpainting model to capture a long term relationship among image regions and the distant known regions while while the baseline model tends to be short-sighted which propagates the most nearby features. For instance, in the second example, both the model with CA layer and CR loss correctly capture the grid pattern while the baseline model just extends the lines. 
The first two rows of Table~\ref{table_metric_abla1} show the quantitative comparison of the baseline model (Baseline) and the baseline model with the CR loss (Baseline+CR). 
The last two rows show the comparison of the original DeepFillv2 and the model obtained by removing CA layer and applying the CR loss (DeepFillv2-CA+CR) with network architectures and training protocols unchanged. From this table, we can see that the CR loss can improve the quantitative performance of different models, and DeepFillv2 with CR loss yields better results than the official implementation with CA layer. 
\begin{table}[h]
\setlength{\tabcolsep}{1.5pt}
\caption{\small Comparison with baseline and CA layer on the Places2 validation set. }
\label{table_metric_abla1}
\small
\begin{center}
\begin{tabular}{ccc|ccc|ccc}
\hline
\multicolumn{3}{c|}{\multirow{2}{*}{Method}} & \multicolumn{3}{c|}{Square holes}& \multicolumn{3}{c}{Irregular holes}\\
\multicolumn{3}{c|}{}& L1 error &PSNR &SSIM& L1 error &PSNR &SSIM\\
\hline
\multicolumn{3}{c|}{Baseline} & .0275 & 22.92 & .8386 & .0171 & 27.03 & .8920\\
\multicolumn{3}{c|}{Baseline+CR} & .0263 & 23.36 & .8462 & .0157 & 27.79 & .9002  \\
\hline
\multicolumn{3}{c|}{DeepFillv2 (with CA)} & .0290 & 22.72 & .8429 & .0178 & 27.03 & .8947\\
\multicolumn{3}{c|}{DeepFillv2-CA+CR} & .0262 & 23.37 & 0.8447 &.0158 & 27.77 & .8984\\
\hline
\end{tabular}
\vspace{-10pt}
\end{center}\
\end{table}

\noindent \textbf{Reference patch. }
To quantify the difference of the reference regions used by the CR Loss and CA layer, we measure the average L1 error, PSNR and SSIM of the images constructed by moving each reference patch to the corresponding position in the missing region like piecing together a jigsaw. The quantitative comparison is shown in Table~\ref{table_metric_abla2}, from which we can see that the images pieced by the CR loss have a smaller L1 error, larger PSNR and SSIM, which indicates that the CR loss can find the reference regions closer to the ground-truth. 
\begin{table}[h]
\setlength{\tabcolsep}{1.5pt}
\caption{\small L1 error, PSNR, SSIM of the image pieced with reference patches selected by the CR loss and CA layer on the Places2 validation set with square holes. }
\label{table_metric_abla2}
\small
\begin{center}
\begin{tabular}{ccc|ccc|ccc}
\hline
\multicolumn{3}{c|}{\multirow{2}{*}{Method}} & \multicolumn{3}{c|}{Square holes}& \multicolumn{3}{c}{Irregular holes}\\
\multicolumn{3}{c|}{}& L1 error &PSNR &SSIM& L1 error &PSNR &SSIM\\
\hline
\multicolumn{3}{c|}{DeepFillv2 (with CA)} &.0331 & 21.37 & .8104 & .0232 & 24.39 & .8562\\
\multicolumn{3}{c|}{DeepFillv2-CA+CR} & .0315 & 21.77 & .8135 &.0205 & 25.27 & .8629\\
\multicolumn{3}{c|}{Baseline+CR} & .0310 & 21.90 & .8152 & .0202 & 25.39 & .8643 \\
\hline
\end{tabular}
\vspace{-10pt}
\end{center}
\end{table}

\noindent\textbf{Efficiency. }Since CR loss gets rid of the heavy pair-wise similarity computation in the inference stage, the obtained model is computationally efficient. Table~\ref{table_metric_abla2} shows time complexity and the running time of the networks of the same architecture with the CR loss and CA layer at different resolutions. The efficiency advantages become increasingly evident when the resolution increases. 
\begin{table}[h]
\caption{\small Comparison of time complexity and the running time measured at different resolution. We are not able to evaluate contextual attention layer at $2080\times2048$ on GPU due to limited GPU memory size and report the comparison on CPU instead. $N$ is the number of pixels in the input image. $T_{s}^{GPU}$ and $T_{s}^{CPU}$ represent the running time (in seconds) on GPU and CPU at $s\times s$ resolution, respectively. }
\label{table_metric_abla2}
\small
\begin{center}
\begin{tabular}{cccccccc}
\hline
&Complexity & $T_{512}^{GPU}$ & $T_{1024}^{GPU}$&$T_{2048}^{GPU}$&$T_{2048}^{CPU}$\\
\hline
CA layer & $O(N^2)$ & 0.063 & 0.371 &Fail &266\\
CR loss & $O(N)$ & 0.047 & 0.179 & 0.720 &28.9\\
\hline
\end{tabular}
\vspace{-20pt}
\end{center}
\end{table}
\section{Conclusion \& future work}
In this paper, we proposed to train a generative image inpainting model with an auxiliary contextual reconstruction branch. 
A learnable contextual reconstruction loss (CR loss) is designed to jointly optimize a reference-based reconstructor with the inpainting generator to teach the path-borrowing behavior to an attention-free generator. It brings no extra computation or parameters to the model and can be applied to almost any network architecture. Experiments show that our inpainting model with the proposed CR loss compares favourably against state-of-the-art in terms of quantitative measurements and visual quality. CR loss can be extended for distilling more information to an inpainting model through the reconstruction of higher-level representation of images,~\eg semantic layout. This might be an interesting topic for future work.

{\small
\bibliographystyle{ieee_fullname}
\bibliography{egbib}
}

\end{document}